# Template-based matching using weight maps

Kwie Min Wong

## 1 Introduction

In this report we try to improve the detection rate of template matching [1-8] assigning higher weights to certain area's of the templates.

The template matcher uses the normalized correlation coefficient (NCC) as the distance between the template and the face image area. The NCC formula is given below).

$$r = \frac{\sum_{i=1}^{n}(X_i - \bar{X})(Y_i - \bar{Y})}{\sqrt{\sum_{i=1}^{n}(X_i - \bar{X})^2}\sqrt{\sum_{i=1}^{n}(Y_i - \bar{Y})^2}}$$

In this formula $X_i$ is a pixel on the template and $Y_i$ is a pixel on the face image area.

In the next section we try to improve the detection rate by focussing on the iris area. This is done by assigning more weight to the center area of the eye templates when calculating the distance between the eye template and the face area's. The following formula was used to calculate the weighted NCC value:

$$r = \frac{\sum_{i=1}^{n} W_i(X_i - \bar{X})(Y_i - \bar{Y})}{\sqrt{\sum_{i=1}^{n} W_i(X_i - \bar{X})^2}\sqrt{\sum_{i=1}^{n} W_i(Y_i - \bar{Y})^2}}$$

$W_i$ is the weight of pixel i. Experiments are done with 4 different weight distributions:

1. The uniform distribution: all pixels are equally weighted.
2. Oval shaped gaussian distribution: gaussian distribution where all weights higher than 1 are in the entire eye area.
3. Circular shaped gaussian distribution: gaussian distribution where the weights higher than 1 are in the iris area.
4. Exponential distribution: highest weights are in the iris area.

## 2 Experiments

### 2.1 Test set

Both the training and the test set were taken from the FERET database [2]. The training set was used to create 80 left and right eye templates (160 total). This was done by hand and may contain small human errors. The test set contains 50 faces (face images include ears, hair, background, etc).

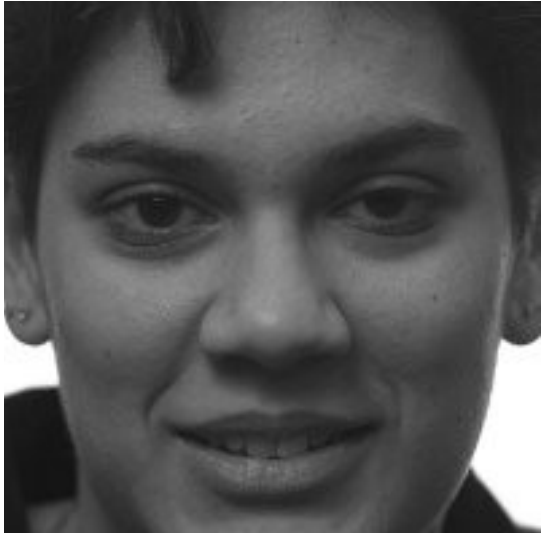

Illustration 1: Example of a test image

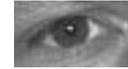

Illustration 2: Example of a right eye template

## 2.2 Uniform distribution

In the uniform distribution all pixel locations have a weight of 1.

|           | 10  | 45  | 80  |
|-----------|-----|-----|-----|
| Right eye | 80% | 88% | 94% |
| Left eye  | 76% | 78% | 88% |

Table 1: Detection percentage (error <8 pixels)

Table 1 shows the percentage of the eyes that were detected location is less than 8 pixels away from the real location. The labels of the columns are the amount of templates used.

Here we see that for both left and right eyes the amount of correct detected eye locations improve when more templates are used.

## 2.3 Gaussian distribution

### 2.3.1 Ellipse shaped

In the gaussian distribution the pixels in the center have a higher weight than the ones on the border. In this experiment we use a weight map where all weights above 1 are inside an ellipse region. The pixels in the eye have higher weights than the background pixels.

The following function was used to generate the weight map:

$$f(x,y) = A e^{-(\frac{(x-x_o)^2}{2\sigma^2} + \frac{(y-y_o)^2}{2\sigma^2})}$$

Because all templates vary in size, we used the values that matched the average template size best (average size is 44x22 pixels): A = 5 (amplitude), $\sigma_x$ = 16, $\sigma_y$ = 8.

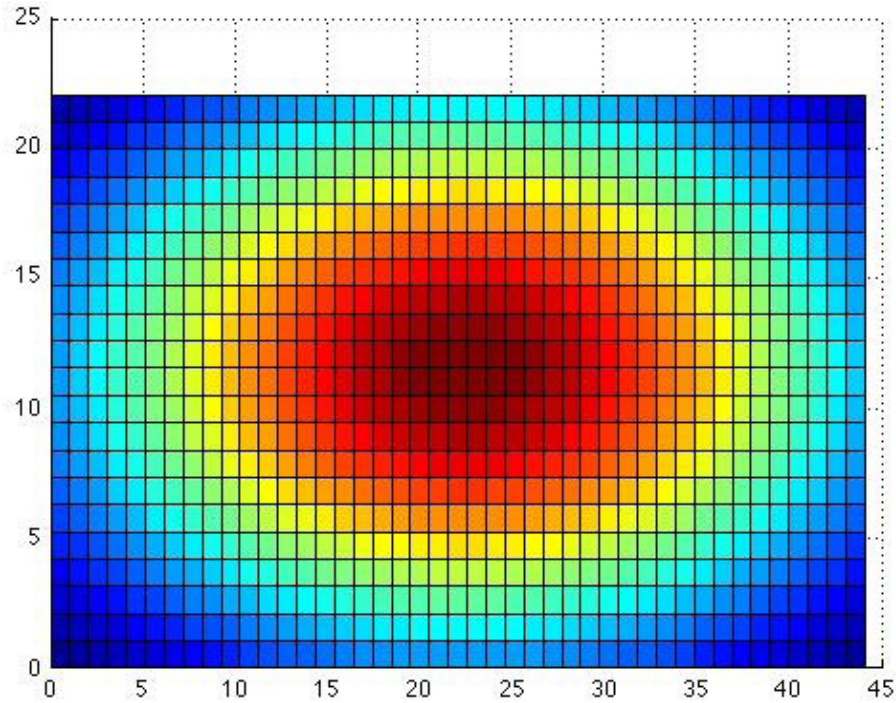

*Illustration 3: Ellipse shaped gaussian weight map*

Illustration 3 is an example of an ellipse shaped gaussian weight map. Red regions have a value of 5 and deep blue regions a value of 1.

The detection rates are shown in Table 2.

|           | 10  | 45  | 80  |
|-----------|-----|-----|-----|
| Right eye | 86% | 92% | 90% |
| Left eye  | 94% | 82% | 88% |

*Table 2: Detection percentage (error < 8 pixels) a gaussian distribution with values A=5, $\sigma_x$ = 16 and $\sigma_y$ = 8*

|            | 10    | 45   | 80   |
|------------|-------|------|------|
| Right eye  | +6%   | +4%  | -4%  |
| Left eye   | +18%  | +4%  | +0%  |

*Table 3: Detection rate increase when using a (elliptical) gaussian weight distribution compared to uniform weight distribution*

When using the gaussian distribution the detection rate improves depending on the amount of templates used. In Table 3 the results of the gaussian map is compared to the results of the uniform map. Here we see that when using 10 templates the increase of the detection rate is much larger then when using 80 templates. For the right eye the detection rate even decreases by 4% when using 80 templates.

2.3.2 Circular

Using the gaussian function from the previous section with values $A = 5$, $\sigma_x = 8$ and $\sigma_y = 8$, we get a weight map with highest weights in the center.

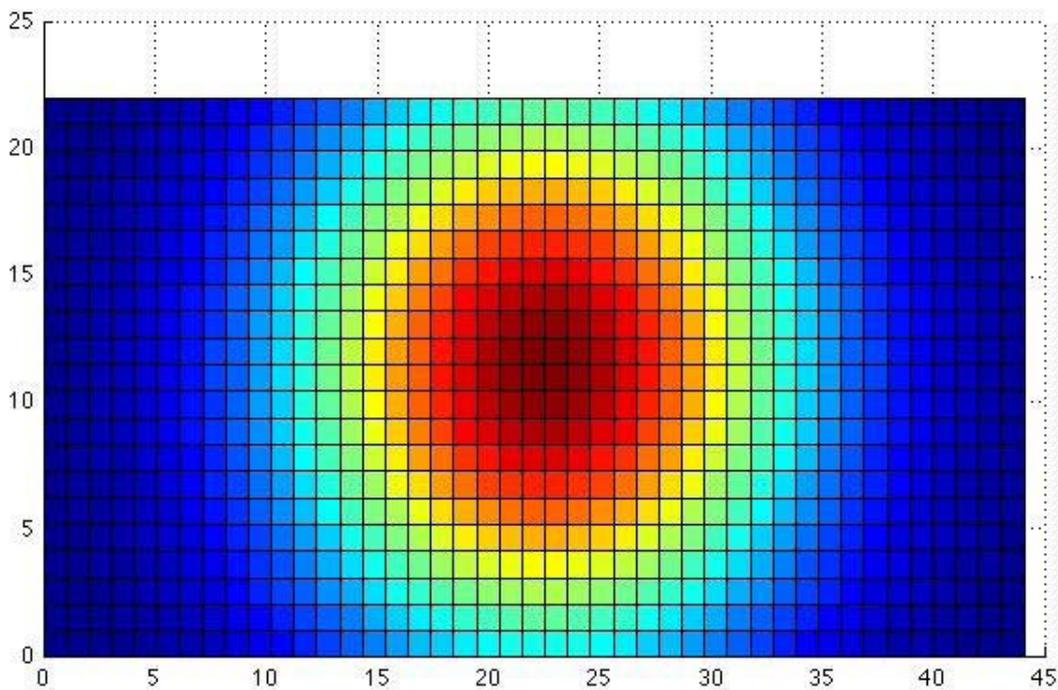

*Illustration 4: Circular gaussian weight mask*

The idea is that the iris is the most important part of the eye. The results of using this circular gaussian weight distribution are listed in Table 4.

|           | 10   | 45   | 80   |
|-----------|------|------|------|
| Right eye | 90%  | 90%  | 94%  |
| Left eye  | 92%  | 78%  | 88%  |

Table 4: Detection percentage (error < 8 pixels). Generated by using a gaussian distribution with values A=5, $\sigma_x$ = 8 and $\sigma_y$ = 8

|           | 10    | 45   | 80   |
|-----------|-------|------|------|
| Right eye | +10%  | +2%  | +0%  |
| Left eye  | +16%  | +0%  | +0%  |

Table 5: detection rate increase when using a circular gaussian weight distribution compared to the uniform distribution

In Table 5 the results of the circular gaussian distribution is compared the uniform distribution. Here we see that like the ellipse shaped gaussian distribution, the detection rate increase, become less when the number of templates are increased. When using less templates, gaussian weight distribution give better results than uniform distribution. When using more templates, the detection rates are about the same.

## 2.4 Exponential distribution

The following formula was used to generate an exponential weight map:

$$f(x,y) = A\, e^{\left|\frac{x-x_o}{b} + \frac{y-y_o}{c}\right|}$$

The following values were used: A = 5, b = 10 and c = 10.

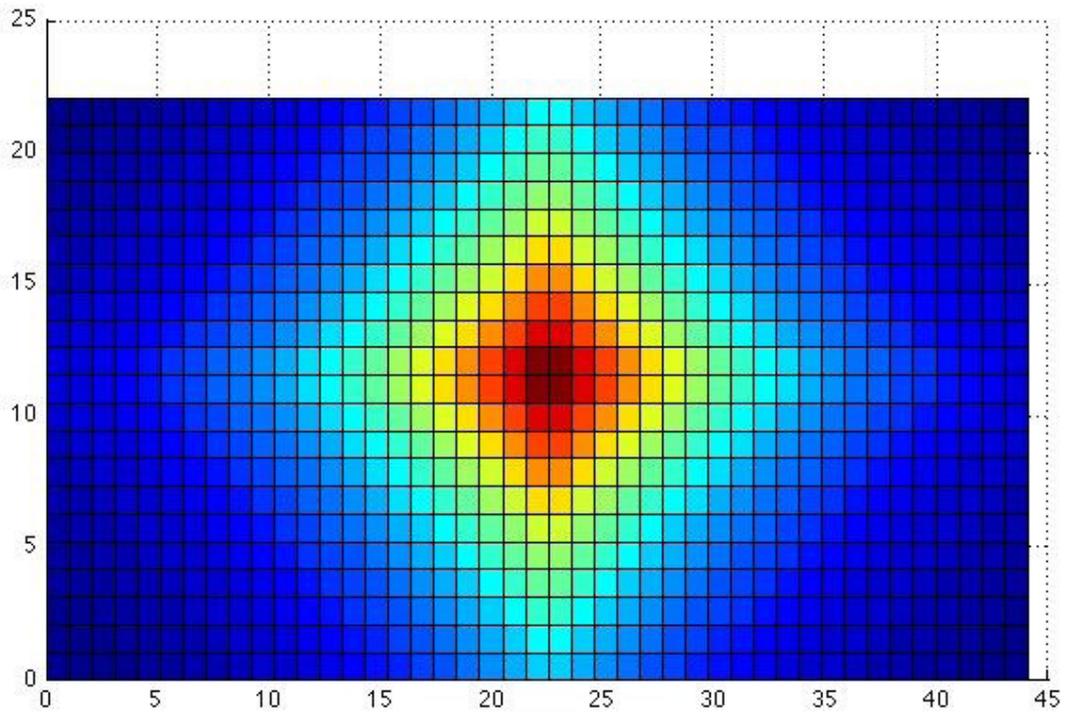

*Illustration 5: Exponential weight mask*

Results are listed in Table 6.

|           | 10  | 45  | 80  |
|-----------|-----|-----|-----|
| Right eye | 92% | 88% | 94% |
| Left eye  | 86% | 80% | 88% |

*Table 6: Detection percentage (error < 8 pixels). Generated by using an exponential weight map.*

|           | 10   | 45  | 80  |
|-----------|------|-----|-----|
| Right eye | +12% | +0% | +0% |
| Left eye  | +10% | +2% | +0% |

*Table 7: Detection rate increase when using a exponential weight map*

In Table 7 the detection rate of the exponential distribution is compared to the results of the uniform distribution. The detection rate increase is similar to the one from the circular gaussian distribution; when using less templates the results are better than the ones from the uniform distribution. The more templates, the lower the detection rate increase.

# 3 Conclusion

In this report we have experimented with three different weight distributions (oval shaped gaussian, circular gaussian and exponential) for eye detection using weighted template matching, and compared the results with the results of the uniform weight distribution. In the uniform distribution all pixel value distances between the template and the image are weighted equally (all have a weight of 1). Results show that when using this weight distribution, the performance of the eye detector depends on the amount of templates; the more templates used, the higher the detection rate.

When we compare the oval shaped gaussian, circular gaussian and exponential distributions with the uniform distribution (Tables 3, 5 and 7), we can see that when using 10 templates the all three weight distributions give better results than the uniform distribution. When we use more templates, then the detection rate of the uniform distribution comes closer to the detection rates of the three other distributions.

If the highest weight values are assigned to the pixel value distances around the iris area, initially the detection rate will be better than when using the uniform weight distribution. Future work will look at using salient points to aid the matching [3].

# Acknowledgements

Portions of the research in this report use the FERET database of facial images collected under the FERET program, sponsored by the DOD Counterdrug Technology Development Program Office.